\documentclass[lettersize,journal]{IEEEtran}
\usepackage{amsmath,amsfonts}
\usepackage{algorithmic}
\usepackage{algorithm}
\usepackage{array}
\usepackage[caption=false,font=normalsize,labelfont=sf,textfont=sf]{subfig}
\usepackage{textcomp}
\usepackage{multirow}
\usepackage{stfloats}
\usepackage{url}
\usepackage{verbatim}
\usepackage{graphicx}
\usepackage{cite}
\usepackage{colortbl}
\usepackage{color}
\usepackage{xcolor}
\usepackage{bbding}  
\usepackage{makecell}
\usepackage{tabularx}
\usepackage{booktabs}
\usepackage[breaklinks,colorlinks]{hyperref}
\usepackage{soul}
\usepackage{bbm}

\soulregister{\ref}7
\soulregister\cite7

\newcommand{\PreserveBackslash}[1]{\let\temp=\\#1\let\\=\temp}

\newcolumntype{C}[1]{>{\PreserveBackslash\centering}p{#1}}
\newcolumntype{R}[1]{>{\PreserveBackslash\raggedleft}p{#1}}
\newcolumntype{L}[1]{>{\PreserveBackslash\raggedright}p{#1}}

\begin{document}
%
\title{WaterMono: Teacher-Guided Anomaly Masking and Enhancement Boosting for Robust Underwater Self-Supervised Monocular Depth Estimation}
%
%
%

\author{Yilin~Ding,
        Kunqian~Li*,~\IEEEmembership{Member,~IEEE},
        Han~Mei,
        Shuaixin~Liu,
        and Guojia Hou,~\IEEEmembership{Member,~IEEE}
\thanks{The research has been supported by the National Natural Science Foundation of China under Grant 62371431, 61901240 and 61906177, in part by the Marine Industry Key Technology Research and Industrialization Demonstration Project of Qingdao under Grant 23-1-3-hygg-20-hy, and in part by the Fundamental Research Funds for the Central Universities under Grants 202262004. (Corresponding author: Kunqian Li)}

\thanks{Yilin Ding, Kunqian Li, Han Mei, and Shuaixin Liu are with the College of Engineering, Ocean University of China, Qingdao 266404, China (dingyilin@stu.ouc.edu.cn; likunqian@ouc.edu.cn; meihan@stu.ouc.edu.cn; liushuaixin@stu.ouc.edu.cn).}
\thanks{Guojia Hou is with the College of Computer Science and Technology, Qingdao University, Qingdao 266071, China (guojiahou@qdu.edu.cn).}

}
%
%

\markboth{}%
{Shell \MakeLowercase{\textit{et al.}}: Bare Dem of IEEEtran.cls for Journals}
%



\maketitle

\begin{abstract}
Depth information serves as a crucial prerequisite for various visual tasks, whether on land or underwater. Recently, self-supervised methods have achieved remarkable performance on several terrestrial benchmarks despite the absence of depth annotations. However, in more challenging underwater scenarios, they encounter numerous brand-new obstacles such as the influence of marine life and degradation of underwater images, which break the assumption of a static scene and bring low-quality images, respectively. Besides, the camera angles of underwater images are more diverse. Fortunately, we have discovered that knowledge distillation presents a promising approach for tackling these challenges. In this paper, we propose WaterMono, a novel framework for depth estimation coupled with image enhancement. It incorporates the following key measures: (1) We present a Teacher-Guided Anomaly Mask to identify dynamic regions within the images; (2) We employ depth information combined with the Underwater Image Formation Model to generate enhanced images, which in turn contribute to the depth estimation task; and (3) We utilize a rotated distillation strategy to enhance the model's rotational robustness. Comprehensive experiments demonstrate the effectiveness of our proposed method for both depth estimation and image enhancement. The source code and pre-trained models are available on the project home page: \url{https://github.com/OUCVisionGroup/WaterMono}.
\end{abstract}


%
\section{Introduction}
\IEEEPARstart{A}{cquiring} depth information is a fundamental task in underwater visual perception, serving as a prerequisite for underwater observation and operation tasks such as visual navigation \cite{Raul2021navigation}, visual reconstruction~\cite{sun2021neuralrecon}, visual localization and mapping~\cite{Liu2024unsupervised}, etc. Additionally, it also plays a crucial role in supporting tasks like underwater image enhancement~\cite{chen2023semantic} and semantic segmentation~\cite{Lu2023depth}. Particularly in the context of underwater image enhancement, depth stands out as one of the most critical parameters within the Underwater Image Formation Model (UIFM)~\cite{akkaynak2018revised}, establishing a natural connection between image enhancement and the task of depth estimation.

The application of underwater depth estimation is primarily carried out by Autonomous Underwater Vehicles (AUVs) or Remotely Operated Vehicles (ROVs). Traditionally, sensors are deployed on them to acquire depth information. There is a range of sensors that can provide depth information, such as sonar~\cite{Trucco2008Devising}, stereo cameras~\cite{williams2014underwater}, and LiDAR~\cite{fu2020lidar}. However, the integration of these sensors into AUVs or ROVs introduces additional costs, weight, and volume. This poses particular challenges for small-sized underwater robots. More importantly, these sensors themselves possess inherent limitations. In the case of sonar, the wide sonar beam leads to insufficient directional resolution, while multiple reflections generate range readings in areas devoid of objects~\cite{kleeman2016sonar}. Regarding stereo cameras, the acquired depth information often contains holes due to low-texture regions and occlusion~\cite{atapour2018comparative}. As for LiDAR, it is hindered by substantial laser scattering in water, which restricts its widespread application~\cite{wu2021defect}.

\begin{figure}[t]
  \centering
   \includegraphics[width=1\linewidth]{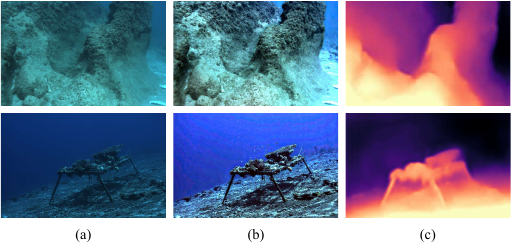}
   \caption{From left to right are the (a) input underwater images, (b) enhanced images and (c) the estimated depth maps obtained by proposed WaterMono. }
   \label{fig:display_outcome}
\end{figure}

Therefore, accurately estimating depth from a single underwater image holds significant appeal and application potential. Some researchers attempt to restore depth information using the UIFM~\cite{drews2016underwater,galdran2015automatic,peng2018generalization,peng2015single}. However, discrepancies between assumed prior conditions and real-world scenes can result in inaccurate outcomes. With the development of deep learning, many researchers have attempted to apply it to underwater monocular depth estimation. Regrettably, progress in this domain utilizing supervised learning approaches has been constrained due to the scarcity of large-scale underwater datasets with ground truth depth values. Initially, researchers attempted to overcome this challenge by employing methods based on Generative Adversarial Networks (GANs)~\cite{gupta2019unsupervised,hambarde2021uw}. They utilized terrestrial datasets in conjunction with generated data to train their models; however, the domain gap between these datasets and authentic underwater images cannot be disregarded, which accounts for the limited generalization of their models.

Recently, the remarkable success achieved in self-supervised monocular depth estimation on land~\cite{garg2016unsupervised,zhou2017unsupervised,godard2017unsupervised,godard2019digging} has motivated an increasing number of researchers to explore its application in underwater scenarios~\cite{yang2022underwater,amitai2023self,wang2023underwater}, as it solely relies on a sequence of consecutive frames for training. The efforts made by them have yielded certain accomplishments; however, they have not conducted a systematic analysis of the impacts posed by the unique underwater environment on self-supervised depth estimation and addressed these challenges in a targeted manner. According to our observation and analysis, the challenges encountered in self-supervised depth estimation tasks underwater can be summarized as follows: (1) the presence of unique dynamic regions underwater, such as rippling caustics and marine organisms, which disrupt the assumption of illumination consistency and a static scene; (2) the inevitable and intricate degradation of underwater images; (3) underwater robots are capable of capturing diverse perspectives of underwater images, thereby imposing higher demands on the model's rotational robustness.

To address these three problems, in this paper we propose an underwater self-supervised framework for monocular depth estimation based on knowledge distillation~\cite{hinton2015distilling} and with enhanced images as a byproduct. The sample results of the presented method for depth estimation and image enhancement are showcased in Fig. \ref{fig:display_outcome}. In summary, our contributions can be succinctly summarized as fourfold:
\begin{itemize}
    \item We propose a novel framework for knowledge distillation learning and utilize a Teacher-Guided Anomaly Mask (TGAM) to address the dynamic regions that disrupt training.
    
    \item We exploit the Underwater Image Formation Model (UIFM) to enhance images and achieve improved depth estimation performance along with enhanced images displaying high visual quality.
    
    \item To enhance the rotational robustness of our model, we introduce a rotated distillation strategy.
    
    \item Through extensive experiments on the challenging FLSea dataset~\cite{randall2023flsea}, we demonstrate that our method outperforms the existing state-of-the-art approaches.
\end{itemize}

Note that we are not the first to combine underwater self-supervised depth estimation with underwater image enhancement (Wang et al., 2023 \cite{wang2023underwater}; Varghese et al., 2023\cite{varghese2023self}). However, previous efforts have simply treated enhancement as a subsequent task following depth estimation. In contrast, we leverage the results of enhancement to assist depth estimation in turn, establishing a mutually beneficial relationship between the two tasks and integrating them reciprocally.

\section{Related Works}\label{sec:Rel}
\subsection{Self-Supervised Monocular Depth Estimation}
To address the challenge of gathering ground-truth data, Garg et al.\cite{garg2016unsupervised} achieved self-supervised depth estimation by transforming it into a novel view synthesis problem and minimizing the photometric loss between the left image and the synthesized right image, even though they still required stereo images for training. Zhou et al. \cite{zhou2017unsupervised} additionally trained a pose network to estimate the camera motion between two sequential frames, enabling self-supervised training using monocular sequential images. The Monodepth2 framework, introduced by Godard et al.\cite{godard2019digging}, addresses occlusion more effectively by incorporating a minimum loss strategy between preceding and following frames and implementing auto-masking (AM) to filter out moving objects. Monodepth2 has emerged as the most widely adopted baseline in this field. The HR-Depth model \cite{lyu2021hr} is built upon Monodepth2, with a redesigned skip-connection to enhance the extraction of high-resolution features. Recently, Zhang et al.\cite{zhang2023lite} proposed Lite-Mono, a hybrid architecture that efficiently combines CNNs and Transformers. Its lightweight design renders it suitable for deployment on devices such as AUVs. However, most existing self-supervised depth estimation methods tend to overfit the vertical position of pixels, neglecting other valuable information such as scene content and structure. As a result, inaccurate depth predictions can occur in underwater scenarios where the AUV may operate at significant tilt angles or even upside down.

Some works have also attempted to improve the results of self-supervised monocular depth estimation using knowledge distillation~\cite{hinton2015distilling}. Ren et al.~\cite{ren2022adaptive} proposed to select the best value from the depth predictions of multiple depth decoders to assist in training the student network. Petrovai and Nedevschi\cite{petrovai2022exploiting} proposed to rescale the depth predictions of the teacher network and filter them using 3D consistency check to generate high-quality pseudo labels for the student network to learn from. However, these knowledge distillation methods do not deviate from the traditional framework, failing to fully utilize pseudo labels.

\subsection{Underwater Depth Estimation}
The absorption and scattering of light by water present significant challenges for estimating underwater depth. However, these phenomena also provide certain cues for the depth estimation task because the appearance of underwater objects varies based on their distance from the camera. Based on this, traditional methods for underwater depth estimation have been proposed to restore scene appearance and estimate depth as a by-product. Many researchers choose to apply the dark channel prior (DCP)\cite{he2010single} and adapt it for underwater environments \cite{drews2016underwater, galdran2015automatic, peng2018generalization}. Additionally, Peng et al. \cite{peng2015single} proposed the use of underwater image blurriness to estimate depth maps, while Song et al. \cite{song2018rapid} developed a depth estimation model based on the underwater light attenuation prior (ULAP), which allows for effective estimation of background light and transmission maps. However, due to their reliance on fragile priors, the effectiveness of these methods cannot be guaranteed.

With the advancement of deep learning (DL), researchers have attempted to estimate the depth of underwater images by using DL-based methods, while the scarcity of large-scale underwater datasets with accurate depth annotation has always plagued these approaches. One approach to tackle this issue involves the utilization of GAN-based methodologies. For instance, UW-Net\cite{gupta2019unsupervised} utilizes the principles of cycle-consistent learning to acquire mapping functions that connect unpaired RGB-D terrestrial images with diverse underwater images, thus only requiring terrestrial depth ground truth. Similarly, UW-GAN \cite{hambarde2021uw} employs a GAN for depth generation, guided by supervision from a synthetic underwater dataset. Another perspective is to utilize weakly-supervised strategies like WsUID-Net\cite{li2024learning}, which uses manually annotated depth trendlines instead of dense depth maps to train the network.

Recently, the community has recognized that self-supervised depth estimation may represent a more promising solution. Yang et al.\cite{yang2022underwater} achieved the pioneering accomplishment of underwater self-supervised depth estimation by collecting underwater videos themselves and proposed using an optical flow estimation network to handle occlusions. Randall and Treibitz proposed FLSea\cite{randall2023flsea}, an underwater visual dataset comprising consecutive frames with ground truth depth information. This initiative has facilitated the advancement of self-supervised depth estimation in underwater environments. Amitai and Treibitz\cite{amitai2023self} examined how the reprojection loss changes underwater using the FLSea dataset. Wang et al.\cite{wang2023underwater} proposed an iterative pose network and suggested enhancing underwater images by utilizing the results of self-supervised depth estimation, but they solely consider image enhancement as a downstream task of depth estimation without fully exploring the intricate relationship between these two tasks.

\subsection{Underwater Image Enhancement}
The methods for enhancing underwater images can be broadly categorized into traditional approaches and deep learning-based techniques. The traditional methods are further divided into two categories: model-free approaches and model-based approaches. Model-free methods aim to improve the visual quality of images without explicitly modeling the degradation process. Such methods primarily include Rayleigh-stretching\cite{ghani2014underwater}, color adjustment \cite{iqbal2010enhancing}, Retinex algorithms \cite{fu2014retinex,zhang2017underwater}, etc. Model-based methods utilize UIFM to restore degraded images \cite{Hou2024Non}. The most prevalent model includes various variants of DCP \cite{drews2016underwater, galdran2015automatic, peng2018generalization}. Another widely applied model is the Jaffe-McGlamery model\cite{mcglamery1980computer}. Recently, Akkayanak and Treibitz proposed a more accurate physical imaging model\cite{akkaynak2018revised}, which provides a precise estimation of the attenuation coefficient. They further introduced the Sea-thru method\cite{akkaynak2019sea} that utilizes RGB-D images to recover colors in underwater scenes.

The task of enhancing underwater images remains a challenge task for deep learning-based approaches, primarily due to the inherent difficulty, if not impossibility, of obtaining ground truth (i.e. the corresponding clear image) for degraded images. Initially, researchers utilized GAN-based methods \cite{guo2019underwater, li2018emerging, li2017watergan} to tackle this challenge. Subsequently, Li et al. \cite{li2019underwater} introduced the UIEB dataset by manually selecting optimal candidates from various enhancement techniques as ground truth. This led to a series of supervised deep learning-based approaches centered around the UIEB dataset, such as Water-Net\cite{li2019underwater}, Ucolor\cite{li2021underwater}, TPENet\cite{TPENet}, and TCTL-Net \cite{li2023tctl}. Unfortunately, their enhancement references used for training are not absolute truth values, and it is still challenging for data-driven strategies to obtain realistic and reliable enhancement results. Recently, researchers have increasingly focused on the joint optimization and resolution of depth estimation tasks in enhancement tasks \cite{Ye2020Joint}. However, the incorporation of enhancement tasks to assist depth estimation remains limited.

\section{Preliminaries and Motivation}\label{sec:Pre}

\begin{figure}[t]
  \centering
   \includegraphics[width=0.9\linewidth]{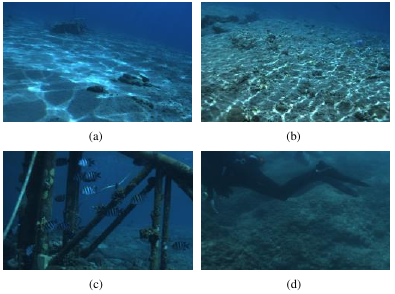}
   \caption{Examples of dynamic regions in the FLSea dataset. (a)\&(b) caustics, (c) fish, (d) diver.}
   \label{fig:anomalous_regions}
\end{figure}

\subsection{Self-Supervised Training for Depth Estimation}
Self-supervised monocular depth estimation method involves the simultaneous training of both a depth network and a pose network  \cite{zhou2017unsupervised,godard2019digging}. Assuming that the target frame $I_{t}$ and the source frames $I_s\in\{I_{t-1},I_{t+1}\}$ are consecutive frames sampled from a video, the depth network estimates the corresponding depth map $\widehat{D}_{t}$ based on the input $I_{t}$. Simultaneously, the pose network estimates the inter-frame pose change $\widehat{T}_{t\to s}$ using inputs $I_{t}$ and $I_{s}$. If the intrinsic camera matrix $K$ is known, we can establish the pixel-to-pixel correspondence between any point $p_{s}$ in frame $I_{s}$ and another point $p_{t}$ in frame $I_{t}$ through the following procedure:
\begin{equation}
\label{correspondence}
p_s\sim K\widehat{T}_{t\rightarrow s}\widehat{D}_t(p_t)K^{-1}p_t.
\end{equation}

Then, a reconstructed target frame $\hat{I}_{t}$ can be generated based on this correspondence, and the network can be optimized with respect to the photometric error loss. The photometric error $pe$ between $I_{t}$ and $\hat{I}_t$, as commonly formulated following \cite{godard2017unsupervised}, typically combines the use of $L1$ norm and Structural Similarity Index (SSIM) \cite{wang2004image}:
\begin{equation}
\label{photometric_loss}
pe(I_t,\hat{I}_t)=\frac\alpha2{\left(1-SSIM(I_t,\hat{I}_t)\right)}+(1-\alpha){\left\|I_t-\hat{I}_t\right\|}_1,
\end{equation}
where $\mathrm{\alpha=0.85}$. \cite{godard2019digging} proposed addressing occlusions by exclusively selecting the minimum pixel-wise reprojection error. The final photometric loss $L_{p}$ is therefore
\begin{equation}
\label{final_photometric_loss}
L_p(I_t,\hat{I}_t)=\min_{I_s}pe(I_t,\hat{I}_t).
\end{equation}

Besides, edge-aware smoothness\cite{godard2017unsupervised} is also commonly used, which is a broad consensus in this task:
\begin{equation}
\label{edge_aware_smoothness}
L_s=|\partial_x\widehat{D}_t|e^{-|\partial_xI_t|}+|\partial_y\widehat{D}_t|e^{-|\partial_yI_t|},
\end{equation}
where $\partial_{x}$ and $\partial_{y}$ are image gradient along horizontal and vertical axes, respectively.

\subsection{Underwater Image Characteristics and the Challenges}

\subsubsection{Unique Dynamic Regions}
Self-supervised monocular training assumes that all objects in the scene are stationary, but it is evident that dynamic regions are prevalent underwater. Examples include rippling caustics, marine life, and diver, as shown in Fig. \ref{fig:anomalous_regions}. These regions can disturb the reprojection process during training, leading to anomalous losses, consequently compromising the performance of the network.

The caustics refer to the intricate physical phenomena that arise from the refraction of light rays projected onto the undulating surface of water\cite{agrafiotis2023seafloor}. This creates illumination patterns on the seabed, which vary in time due to dynamic surface waves. Furthermore, the aquatic environment harbors a plethora of marine organisms, predominantly fish, showcasing diverse morphologies and an extensive array of species. Moving marine organisms also constitute dynamic regions. Finally, underwater images are usually captured by divers or underwater robots, and their presence is often portrayed in the frames, creating an additional type of dynamic region.

Filtering out dynamic regions or moving objects has long been a key focus in self-supervised depth estimation. The most classic approach is auto-masking (AM)\cite{godard2019digging}, which, however, can only filter out objects moving at the same velocity as the camera. It proves ineffective when confronted with dynamic regions exhibiting more complex motion patterns. Some other approaches employ semantic segmentation network or optical flow estimation network to identify moving 
 objects\cite{wimbauer2021monorec,shen2023dna}, which require additional semantic labels or network structures.

\subsubsection{Degradation of Underwater Images}
The degradation of underwater images is primarily caused by the absorption and scattering of light in the water medium\cite{li2022beyond}. The propagation of light in water leads to energy loss due to its absorption, resulting in a color bias observed in underwater images. This absorption process is selective\cite{schettini2010underwater}, with red light of longer wavelengths experiencing the most rapid attenuation in underwater environments. Consequently, this phenomenon gives rise to a predominant blue and green color bias in underwater images.

\begin{figure*}[t]
\centering
	\includegraphics[width=1\linewidth]{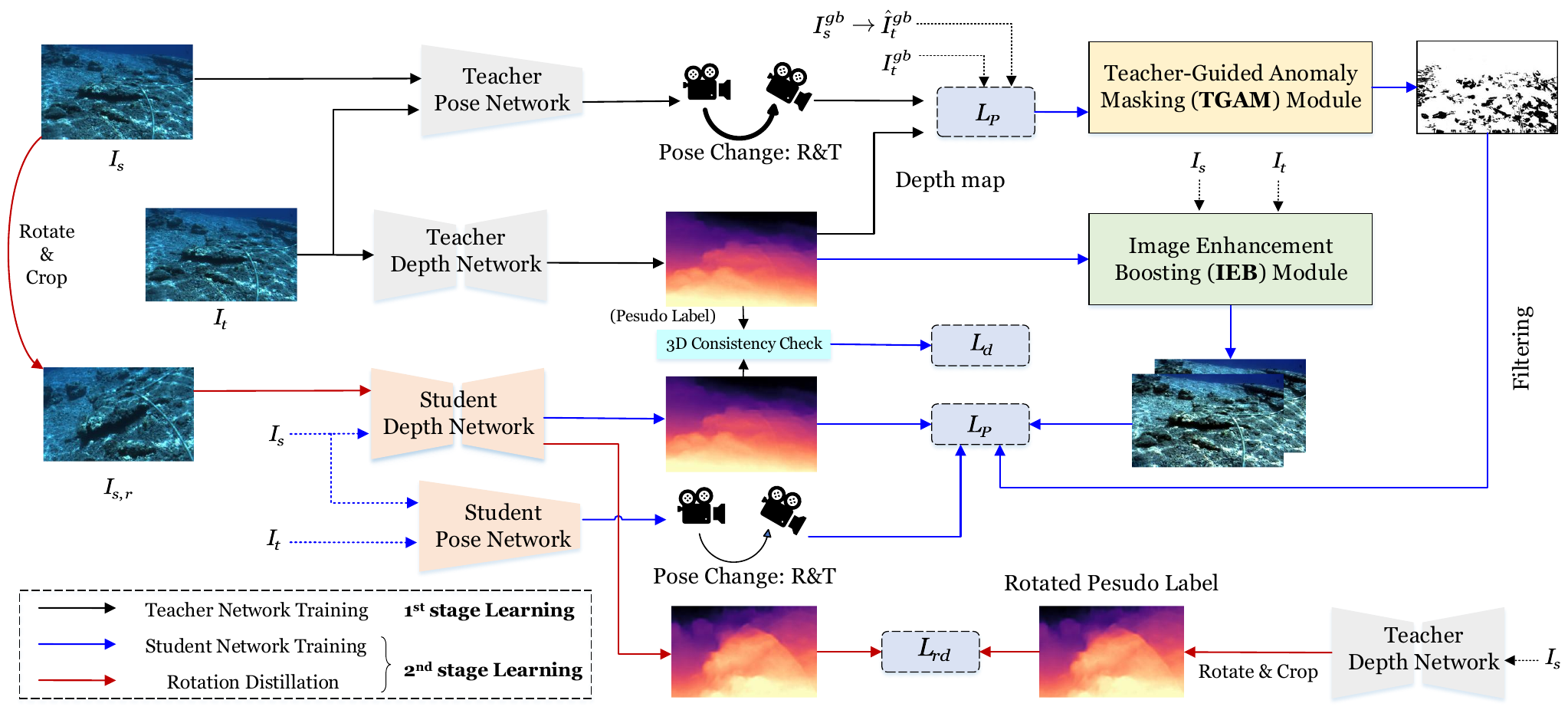}
    \caption{Overview of our WaterMono training pipeline. In the first stage, we conduct self-supervised training for a teacher depth network and a teacher pose network. Using the teacher depth network, we generate pseudo depth labels for input images. In the second stage, we freeze the teacher networks and train the student depth and pose networks from scratch. The training approach for the student networks involves a mixture of self-supervised and supervised techniques. The student networks compute photometric loss on enhanced images and utilizes TGAM to filter out dynamic regions. Pseudo labels, filtered through 3D consistency check, are also employed to supervise the student depth network. Additionally, paired images and pseudo labels under rotated camera angles are generated through rotation transformations for the depth student network's supervised learning.}
    \label{fig:pipeline}
    \vspace{-1em}
\end{figure*}

The scattering of light by water can be categorized into two distinct types: forward scattering and backward scattering\cite{raveendran2021underwater}. Forward scattering refers to the phenomenon where light reflected by the target object deviates due to the influence of suspended particles in the water during its projection onto the camera, resulting in image blurring. Backward scattering represents the phenomenon where natural light entering the water is scattered towards the camera due to suspended particles, causing low contrast and a hazy effect on images\cite{qi2022sguie,raveendran2021underwater}. Among these detrimental factors, blurring and low contrast have a particularly severe impact on self-supervised depth estimation tasks as they lead to either minimal or incorrect photometric losses in regions with significant degradation, consequently hindering proper optimization direction for neural networks.

\subsubsection{Diverse and Dynamically Changing Camera Angles}
The motion of vehicles on land is typically confined to a two-dimensional plane, whereas underwater robots possess six degrees of freedom and navigate in three-dimensional space, allowing them to move freely in water and observe the underwater environment from various perspectives and angles. As a result, images captured by underwater robots exhibit a wider range of camera angles. We collectively refer to camera angles other than the frontal view as rotated camera angles. This feature places a higher demand on the rotational robustness of underwater depth estimation methods.

\section{Proposed Method}\label{sec:Pro}

\subsection{Overall Framework of WaterMono}
Combining the aforementioned characteristics of underwater images with the challenges posed by depth estimation tasks, as depicted in Fig. \ref{fig:pipeline}, we propose a two-stage training pipeline named WaterMono for underwater monocular depth estimation, which also yields enhanced images as a by-product. In the first stage, the teacher networks are trained in a self-supervised manner on consecutive images with the photometric loss. The teacher depth network is then utilized to infer depth across the entire training set in order to generate pseudo labels.

In the second stage, to mitigate the interference of underwater dynamic regions on self-supervised training, we reconstruct images using pseudo labels to generate masks for these regions, thereby excluding interference. To address degradation in underwater images, we enhance them using Image Enhancement Boosting (IEB) module to obtain more reasonable photometric losses. Additionally, to overcome the instability of self-supervised training, pseudo labels will be used to supervise the student depth network after 3D consistency check to remove unreliable depth values. Furthermore, to cope with the diverse camera angles underwater, we propose a rotated distillation strategy, creating pairs of samples under rotated camera angles by rotating images and pseudo labels.

The second stage has some key points to note: if the input is the original image, the student depth network will be constrained by both the supervised loss from pseudo labels and the self-supervised photometric loss. Unlike the first stage, in this case, the photometric loss is calculated on the enhanced image and filtered by TGAM to remove interference. If the input is a rotated image, then only the supervised loss from pseudo labels constrains the student depth network.

The foundation of our work is built upon Lite-Mono\cite{zhang2023lite}. Both the teacher and student networks adopt the unaltered Lite-Mono network architecture.

\subsection{Teacher-Guided Anomaly Masking}
The training of a self-supervised monocular depth estimation network relies on the assumption of a stationary scene, but dynamic regions disrupt this assumption and result in anomalous losses. Researchers have attempted various methods to localize dynamic regions, thereby eliminating the corresponding anomalous losses\cite{casser2019depth, wimbauer2021monorec}. However, the observation reveals that after a few epochs of training (e.g., 30 epochs), the network becomes proficient in predicting relatively accurate overall depth values. By this time, the losses incurred by dynamic regions obviously outweigh those caused by imprecise depth estimation. This discovery offers a fresh perspective, enabling us to exploit a somewhat well-trained depth estimation network and pose estimation network for localizing moving objects based on anomalous losses.

Specifically, we first train a depth network and a pose network through self-supervised learning, and then freeze their weights to use them as the teacher networks. Then we input the target frame $I_{t}$ and the source frames $I_{s}$ into the teacher networks. Based on their estimated depth and pose changes, we obtain the reconstructed target frames $\hat{I}_t$. We calculate the photometric loss which is formulated as
\begin{equation}
\label{blurred_Lpe}
L_p(I_t^{gb},\hat{I}_t^{gb})=\min_{I_s\in\{I_{t-1},I_{t+1}\}}pe(I_t^{gb},\hat{I}_t^{gb}).
\end{equation}
where $I_t^{gb}$ and $\hat{I}_t^{gb}$ are the Gaussian blurred version of $I_{t}$ and $\hat{I}_t$ to reduce the influence accumulated tiny errors. Specifically, 
\begin{equation}
\label{gaussian_blur_1}
I_t^{gb}=\frac1{\sqrt{2\pi}\sigma}\sum_{i=-k}^k\sum_{j=-k}^kI_t\left(x+i,y+i\right)e^{-\frac{i^2+j^2}{2\sigma^2}},
\end{equation}
\begin{equation}
\label{gaussian_blur_2}
\hat{I}_{t}^{gb}=\frac{1}{\sqrt{2\pi}\sigma}\sum_{i=-k}^{k}\sum_{j=-k}^{k}\hat{I}_{t}\left(x+i,y+i\right)e^{-\frac{i^{2}+j^{2}}{2\sigma^{2}}}.
\end{equation}
where $k$ and $\sigma$ are the window size and standard deviation of the adopted  Gaussian blur kernel, respectively.

Assuming that the teacher network has received effective training to a certain extent, if the assumption of a static scene holds true, there should not be significant photometric differences between $I_{t}$ and $\hat{I}_t$. Otherwise, regions with significantly larger losses can be considered as dynamic regions. Properly masking these areas during the training of the student networks is beneficial for minimizing interference and achieving better results than the teacher network.

Therefore, we introduce Teacher-Guided Anomaly Masking (TGAM) to facilitate the removal of dynamic regions. Let $\left[ \cdot \right] $ denote the Iverson bracket, and $T(i)$ represents the masking threshold for the $i$th image, TGAM produces a mask $m_{t}$ by
\begin{equation}
\label{TGAM}
m_t=[L_p(I_t^{gb},\hat{l}_t^{gb})<T(i)].
\end{equation}

\begin{figure}[t]
  \centering
   \includegraphics[width=0.95\linewidth]{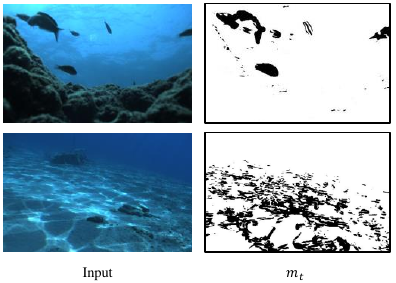}
   \caption{Examples of $m_{t}$, where black pixels are removed from loss. We can see that, $m_{t}$ can mask dynamic regions like fish and caustics.}
   \label{fig:TGAM_examples}
\end{figure}

Inspired by\cite{wang2021regularizing}, we adopt a statistical strategy similar to exponential moving weighted average to determine the threshold $T(i)$. For the photometric loss $L_p$ of each image in the teacher networks, we take its $\epsilon$-th percentile \(t(i) = p(L_p, \epsilon)\). Then we calculate $T(i)$ using the Eq. (\ref{Threshold}):
\begin{equation}
\label{Threshold}
T(i)=\beta\times T(i-1)+(1-\beta)\times t(i),
\end{equation}
where the momentum parameter $\beta$ is set to to $0.98$, and $\epsilon$ is set to $5$. With these parameter settings, TGAM is designed to eliminate the top $5\%$ of pixels with the highest $L_p$ across the entire dataset, while allowing for some flexibility in fluctuations between different images. This strategy is simple yet effective. As illustrated in Fig. \ref{fig:TGAM_examples}, a collection of TGAM examples vividly demonstrates its remarkable capability in detecting diverse dynamic regions underwater.

\subsection{Image Enhancement Boosting Module}
The inherent degradation of underwater imaging leads to the blurring and loss of details, which inevitably impedes accurate depth estimation. The incorporation of an image enhancement module is thus a direct and indispensable approach to improve depth estimation. The Akkaynak–Treibitz model\cite{akkaynak2018revised} is presently the most widely embraced underwater imaging formation model, which can be represented as follows:
\begin{equation}
\label{UIFM_1}
I_c=J_ce^{-\beta_c^D(d)*d}+B_c^\infty(1-e^{-\beta_c^B(d)*d}),
\end{equation}
where $c\in\{r,g,b\}$ represents the color channel, $I_{c}$ denotes the degraded image captured underwater, $J_{c}$ refers to the underlying clean image, $\beta_{c}^{D}$ and $\beta_{c}^{B}$ represent the backscatter coefficient and light attenuation coefficient respectively, $B_c^\infty$ is the veiling light, while $d$ represents depth.

In the case of a known value for $d$, as stated in Sea-thru\cite{akkaynak2019sea}, we can estimate $\beta_{c}^{D}$, $\beta_{c}^{B}$, and $B_{c}^\infty$ to acquire the clear image $J_{c}$. However, the degradation parameters obtained through the Sea-thru method exhibit a certain level of randomness, resulting in enhanced images that lack consistency between frames. If we adopt this model for underwater image enhancement, this inconsistency will present challenges for self-supervised depth estimation since inter-frame discrepancies may introduce erroneous photometric losses.

By considering the requirements of inter-frame consistency, we introduce an Image Enhancement Boosting (IEB) module, which is built upon Sea-thru. We try to disregard the influence of distance on backward scattering dependency, thereby proposing a simplified yet more consistent enhancement model for consecutive frames to boost depth estimation. For each degraded image, we treat backward scattering as a constant, simplifying the imaging model to:
\begin{equation}
\label{UIFM_2}
I_c=J_ce^{-\beta_c^D(z)*z}+B_c.
\end{equation}
This enables us to solely estimate $B_{c}$ and $\beta_{c}^{D}$. To estimate $B_{c}$, we perform a search for the darkest 0.1\% pixels in the image and calculate their mean as the backscatter estimation. Regarding $\beta_{c}^{D}$, we assume that degraded images from the same scene, captured at similar times and locations, should possess identical $\beta_{c}^{D}$. Therefore, for each scene, we can extract partial images and jointly estimate $\beta_{c}^{D}$ using the Sea-thru method applied to the entire scene. This ensures inter-frame consistency among images originating from identical scenes.

Furthermore, the incorporation of high-frequency information is indispensable for accurate depth estimation in images, as it plays a pivotal role in comprehending the geometric structure of objects\cite{miangoleh2021boosting,li2021self}. Nevertheless, underwater scenes tend to exhibit increased blurriness in regions with greater depth. To address this issue effectively, we have employed the unsharp masking algorithm and leveraged depth guidance to regulate its sharpening intensity. The depth-weighted sharpening algorithm can be represented as:
\begin{equation}
\label{USM}
I_{en}=(I-\mathcal{F}_L(I))*d^{\prime}+I,
\end{equation}
where the enhanced image is represented by $I_{en}$, the input image by $I$, a low-pass Gaussian filter by $\mathcal{F}_L$, and a weighting coefficient by $d^{\prime}$ which is the normalized estimated depth. By doing so, we are able to enhance degraded details without excessively enhancing clear regions.

It should be noted that the enhanced images will only be used to calculate the photometric loss of the student networks, and the original images will still serve as input to the student networks. By alleviating low contrast and blur in the images, the enhanced images exhibit a more comprehensive and plausible photometric loss compared to the degraded ones, thus better optimizing the model. In doing so, underwater image enhancement transcends from being solely a downstream task of underwater depth estimation to establishing a mutually beneficial relationship where they act as both upstream and downstream tasks for each other.

\subsection{{Selective Distillation}}
To alleviate the instability and local minima problem in self-supervised training, we introduce supervision from pseudo labels. However, pseudo labels from the teacher network are not always reliable. If pseudo labels with significant errors are also involved in knowledge distillation, it can be detrimental to the student network. Therefore, distillation should be selective to exclude unreliable portions of pseudo labels. We achieve this by applying the 3D consistency checks proposed in \cite{petrovai2022exploiting}.

Following\cite{petrovai2022exploiting}, we input both the target frame $I_{t}$ and the source frames $I_{s}$ into the teacher networks to obtain estimated corresponding depths $\widehat{D}_{t}$, $\widehat{D}_{s}$ and the pose change $\widehat{T}_{t\to s}$ between them. Utilizing Equation (\ref{correspondence}), we can obtain the corresponding point $p_{s}$ in $I_{s}$ for any point $p_{t}$ in $I_{t}$. With $\widehat{D}_{s}$, $\widehat{T}_{t\to s}$, and the camera's intrinsic matrix \(K\), we can project $p_{s}$ into the camera coordinate system of the target frame $I_{t}$. Correspondingly, \(p_t\) is also projected into the camera coordinate system of \(I_t\). In the same coordinate system, we measure depth consistency by computing the distance between the coordinates of these two 3D points. Let $\left[ \cdot \right] $ denote Iverson bracket and $\tau$ a predefined threshold that is set to 0.03, a depth consistency mask $m_{c}$ can be generated by
\begin{equation}
\label{mc}
m_c=[\min_{I_s}\lVert\widehat{T}_{t\to s}\widehat{D}_s(p_s)K^{-1}p_s-\widehat{D}_t(p_t)K^{-1}p_t\rVert_1<\tau].
\end{equation}

Using this approach, we assess the consistency of depths at corresponding points between consecutive frames in the pseudo labels. If inconsistencies are observed, we deem them unreliable and unsuitable for distillation into the student depth network. Then, the distillation loss $L_{d}$ can be written as:
\begin{equation}
\label{loss_d}
L_d=\lambda\cdot m_c\cdot\log{(|d_t-d_s|+1)},
\end{equation}
where $d_{t}$ is the pseudo label, and $d_{s}$ is the depth predicted by the depth student network. $\lambda$ is a weight that gradually decays during the training process because as training progresses, the accuracy of student will gradually surpass that of the teacher.

\subsection{Rotated Distillation}
Given that underwater robots have six degrees of freedom, their vision systems usually have a variety of camera angles in free exploration and work application scenarios. However, this phenomenon is not evident in existing underwater depth estimation benchmarks, such as the FLSea dataset. This is primarily because the ROVs must maintain smooth motion and minimal tilt to meet the high-quality requirements for capturing videos used in the depth estimation dataset. Rotating the images in the training set is not a viable solution if we want to improve the model's ability to handle images taken from different camera angles, because self-supervised depth estimation training method heavily relies on pixel correspondence to generate supervision signals through view matching. The introduction of rotation transformation would disrupt this crucial correspondence.

Fortunately, we observe that it is feasible to generate pairs of underwater images and depth pseudo labels at various camera angles by applying rotation transformations to both the input and output of the teacher depth network. In other words, initially, we feed the image from a standard angle into the teacher depth network to acquire the corresponding depth pseudo label. Subsequently, we apply identical rotation transformations to both the images and pseudo labels in order to obtain paired samples. We then input the rotated image into the  student depth network while utilizing the rotated pseudo label as a constraint.

When employing the conventional approach for image rotation, the resulting images may exhibit black borders or empty regions due to alterations in their size, thereby impeding training procedures. Therefore, we perform a center crop after rotating the image by $\theta$ degrees. If the original image size is $[H, W]$, the resulting image size after center cropping is $[h, w]$. The relationship between the two is:
\begin{equation}
\label{center_crop_1}
h=\frac{H\cos\theta-W\sin\theta}{\cos2\theta},
\end{equation}
\begin{equation}
\label{center_crop_2}
w=\frac{W\cos\theta-H\sin\theta}{\cos2\theta}.
\end{equation}

Due to the center cropping that cuts off the pixels and pseudo labels at the corners, the scale of the pseudo labels is disrupted. Therefore, we choose the Pearson correlation coefficient, which is completely unrelated to the scale, as the loss function. The rotated distillation loss $L_{rd}$ is expressed as:
\begin{equation}
\label{L_RD}
L_{rd}=1-\frac{\sum(d_t-\overline{d_t})(d_s-\overline{d_s})}{\sqrt{\sum(d_t-\overline{d_t})^2(d_s-\overline{d_s})^2}},
\end{equation}
where $d_{t}$ and $d_{s}$ are the depths predicted by the teacher depth network and the student depth network, respectively, and $\overline{d_t}$ and $\overline{d_s}$ are their means. When the student depth network's input is a rotated image, it will only be constrained by $L_{rd}$.

The angle $\theta$ at which images and pseudo labels are rotated is a random number between $\mathrm{[-}\gamma\mathrm{,}\gamma\mathrm{]}$, where $\gamma$ refers to the rotation range. $\gamma$ serves as an important hyperparameter; the higher the value of $\gamma$, the greater the model's rotational robustness. We will conduct a detailed analysis of its impact in our ablation study.

\begin{table*}[t]
\caption{Quantitative results of depth estimation on the FLSea dataset~\cite{randall2023flsea} using the OUC split. \textnormal{Best results in each category are in \textbf{bold}. All inputs images are resized to $448\times 288$  and median scaling is applied for all methods. “†” means using a pre-trained model.} }
\centering
\renewcommand{\arraystretch}{1.2}
\begin{tabularx}{\linewidth}{>{\centering\arraybackslash}X|>{\centering\arraybackslash}X|>{\centering\arraybackslash}X|>{\centering\arraybackslash}X>{\centering\arraybackslash}X>{\centering\arraybackslash}X>{\centering\arraybackslash}X>{\centering\arraybackslash}X>{\centering\arraybackslash}X>{\centering\arraybackslash}X>{\centering\arraybackslash}X}
\Xhline{1pt}
\multicolumn{1}{c|}{\multirow{2}{*}{Method}} & \multicolumn{1}{c|}{\multirow{2}{*}{\begin{tabular}[c]{@{}c@{}}Self-\\ Supervised\end{tabular}}} & \multicolumn{1}{c|}{\multirow{2}{*}{Year}} & \multicolumn{4}{c|}{Depth Error($\downarrow$)}                                                                                       & \multicolumn{3}{c|}{Depth Accuracy($\uparrow$)}                                                & \multirow{2}{*}{Params($\downarrow$)} \\ \cline{4-10}
\multicolumn{1}{c|}{}                        & \multicolumn{1}{c|}{}                                                                            & \multicolumn{1}{c|}{}                      & \multicolumn{1}{c}{Abs Rel} & \multicolumn{1}{c}{Sq Rel} & \multicolumn{1}{c}{RMSE} & \multicolumn{1}{c|}{RMSE log} & \multicolumn{1}{c}{$\delta<1.25$} & \multicolumn{1}{c}{$\delta<1.25^{2}$} & \multicolumn{1}{c|}{$\delta<1.25^{3}$}  &                         \\ \hline
\multicolumn{11}{c}{Physical Model-Based}                                                                                                                                                                                                                                                                                                                                                                                            \\ \hline
\multicolumn{1}{c|}{DCP\cite{he2010single}}                    & \multicolumn{1}{c|}{$\times$}                                                                           & \multicolumn{1}{c|}{2009}                  & 0.397                        & 0.901                       & 2.293                     & \multicolumn{1}{c|}{0.527}    & 0.361                     & 0.638                     & \multicolumn{1}{c|}{0.817} & -  \\
\multicolumn{1}{c|}{UDCP\cite{drews2016underwater}}                   & \multicolumn{1}{c|}{$\times$}                                                                           & \multicolumn{1}{c|}{2016}                  & 0.531                        & 1.545                       & 2.494                     & \multicolumn{1}{c|}{0.615}    & 0.295                     & 0.546                     & \multicolumn{1}{c|}{0.728} & -  \\
\multicolumn{1}{c|}{ULAP\cite{song2018rapid}}             & \multicolumn{1}{c|}{$\times$}                                                                           & \multicolumn{1}{c|}{2018}                  & 0.664                        & 2.932                       & 2.971                     & \multicolumn{1}{c|}{0.725}    & 0.257                     & 0.488                     & \multicolumn{1}{c|}{0.667}  & -  \\ \hline
\multicolumn{11}{c}{Deep Learning-Based}                                                                                                                                                                                                                                                                                                                                                                                                        \\ \hline
\multicolumn{1}{c|}{UW-Net†\cite{gupta2019unsupervised}}                  & \multicolumn{1}{c|}{$\times$}                                                                           & \multicolumn{1}{c|}{2019}                  & 1.091                        & 9.310                       & 4.137                     & \multicolumn{1}{c|}{0.920}    & 0.214                     & 0.401                     & \multicolumn{1}{c|}{0.555} & 29.6M                   \\ \hline
\multicolumn{1}{c|}{Monodepth2-Res18\cite{godard2019digging}}        & \multicolumn{1}{c|}{$\checkmark$}                                                                           & \multicolumn{1}{c|}{2019}                  & 0.145                        & 0.229                       & 1.204                     & \multicolumn{1}{c|}{0.205}    & 0.808                     & 0.941                     & \multicolumn{1}{c|}{0.977} & 14.3M\\
\multicolumn{1}{c|}{Monodepth2-Res50\cite{godard2019digging}}        & \multicolumn{1}{c|}{$\checkmark$}                                                                           & \multicolumn{1}{c|}{2019}                  & 0.143                        & 0.292                       & 1.293                     & \multicolumn{1}{c|}{0.200}    & 0.827                     & 0.944                     & \multicolumn{1}{c|}{0.977} & 35.2M \\
\multicolumn{1}{c|}{HR-Depth\cite{lyu2021hr}}                & \multicolumn{1}{c|}{$\checkmark$}                                                                           & \multicolumn{1}{c|}{2021}                  & 0.143                        & 0.259                       & 1.217                     & \multicolumn{1}{c|}{0.198}    & 0.813                     & 0.949                     & \multicolumn{1}{c|}{0.979} & 14.7M \\
\multicolumn{1}{c|}{Lite-HR-Depth\cite{lyu2021hr}}           & \multicolumn{1}{c|}{$\checkmark$}                                                                           & \multicolumn{1}{c|}{2021}                  & 0.182                        & 0.409                       & 1.582                     & \multicolumn{1}{c|}{0.258}    & 0.727                     & 0.909                     & \multicolumn{1}{c|}{0.962} & \textbf{3.1M}\\
\multicolumn{1}{c|}{DIFFNet\cite{zhou2021self}}                 & \multicolumn{1}{c|}{$\checkmark$}                                                                           & \multicolumn{1}{c|}{2021}                  & 0.141                        & 0.256                       & 1.303                     & \multicolumn{1}{c|}{0.208}    & 0.810                     & 0.936                     & \multicolumn{1}{c|}{0.975} & 10.8M \\

\multicolumn{1}{c|}{ManyDepth\cite{watson2021temporal}}                 & \multicolumn{1}{c|}{$\checkmark$}                                                                           & \multicolumn{1}{c|}{2021}                  & 0.148                        & 0.232                       & 1.229                     & \multicolumn{1}{c|}{0.207}    & 0.803                     & 0.939                     & \multicolumn{1}{c|}{0.975} & 26.9M \\

\multicolumn{1}{c|}{MonoViT-small\cite{zhao2022monovit}}           & \multicolumn{1}{c|}{$\checkmark$}                                                                           & \multicolumn{1}{c|}{2022}                  & 0.123                        & 0.172                       & 1.102                     & \multicolumn{1}{c|}{0.180}    & 0.835                     & 0.958                     & \multicolumn{1}{c|}{0.987} & 27.0M\\
\multicolumn{1}{c|}{Lite-Mono\cite{zhang2023lite}}               & \multicolumn{1}{c|}{$\checkmark$}                                                                           & \multicolumn{1}{c|}{2023}                  & 0.116                        & 0.154                       & 1.053                     & \multicolumn{1}{c|}{0.170}    & 0.860                     & 0.964                     & \multicolumn{1}{c|}{0.988} & \textbf{3.1M}\\
\multicolumn{1}{c|}{WaterMono (Ours)}                    & \multicolumn{1}{c|}{$\checkmark$}                                                                           & \multicolumn{1}{c|}{2024}                  & \textbf{0.099}                        & \textbf{0.117}                       & \textbf{0.945 }                    & \multicolumn{1}{c|}{\textbf{0.152}}    & \textbf{0.892}                     & \textbf{0.978}                    & \multicolumn{1}{c|}{\textbf{0.991}} & \textbf{3.1M}  \\ \Xhline{1pt}
\end{tabularx}
\label{tab:contrastive}
\end{table*}

\section{Experiment}
In this section, we initially introduced the specifics of the benchmark dataset and algorithm implementation. Subsequently, through a series of qualitative and quantitative experiments, we showcased that our method can achieve remarkable results in underwater depth estimation while also generating enhanced images with commendable visual quality as an additional outcome. Furthermore, we conducted ablation studies to demonstrate the positive impact of each proposed component on the overall performance of our model. Lastly, we validated the generalization capability of our model.

\begin{figure*}[!ht]
\centering
	\includegraphics[width=0.95\linewidth]{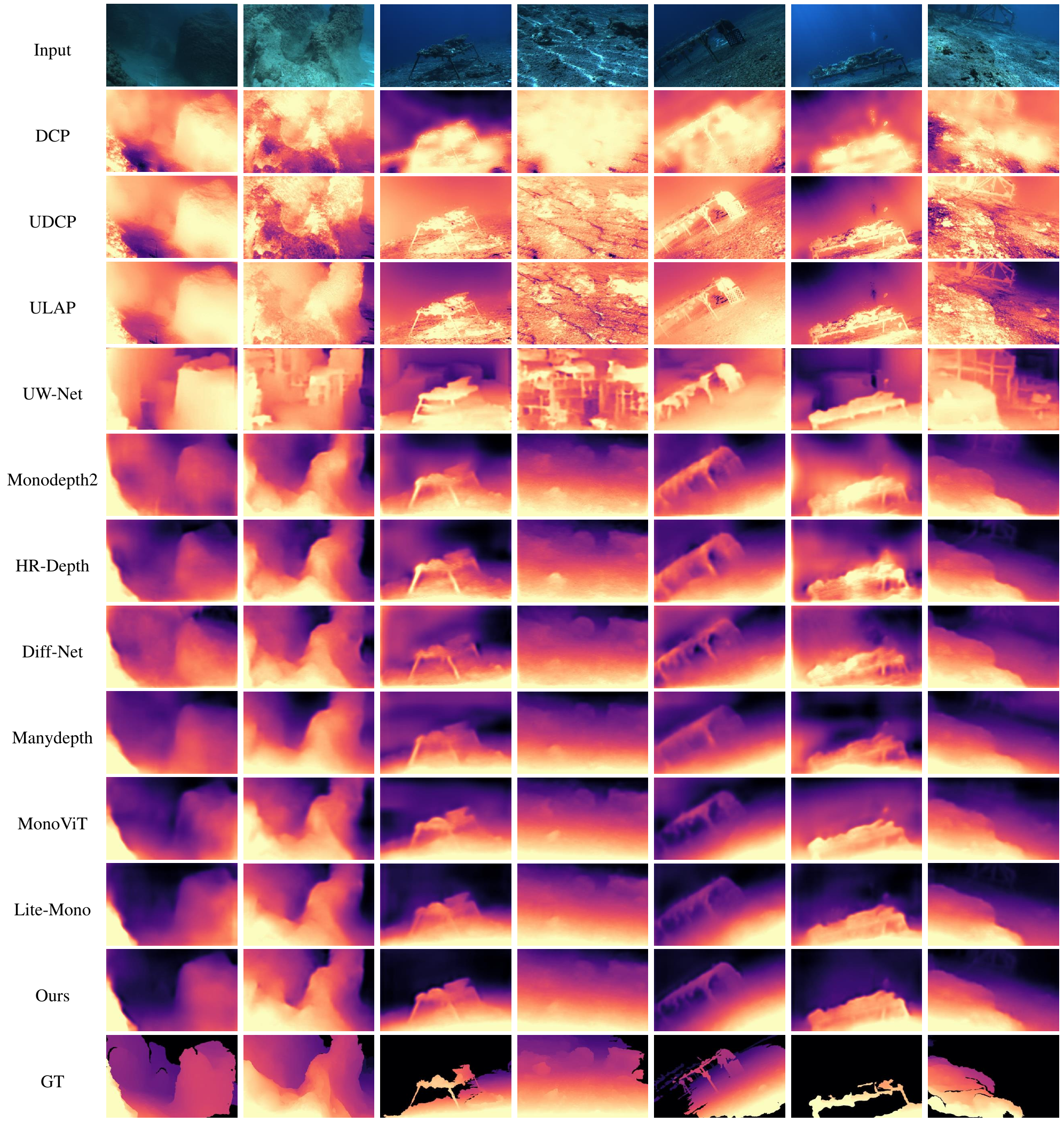}
    \caption{Qualitative comparison on FLSea OUC test set. The first row consists of input images. Results from DCP\cite{he2010single}, UDCP\cite{drews2016underwater}, ULAP\cite{song2018rapid}, UW-Net\cite{gupta2019unsupervised}, Monodepth2\cite{godard2019digging}, HR-Depth\cite{lyu2021hr}, DIFFNet\cite{zhou2021self}, ManyDepth \cite{watson2021temporal}, MonoViT\cite{zhao2022monovit}, Lite-Mono\cite{zhang2023lite} and our method (WaterMono) are listed from the second to the twelfth row. The ground truth is shown at the bottom, where black areas indicate missing depth information in the depth map.}
    \label{fig:test_samples}
    \vspace{-1em}
\end{figure*}

\subsection{Dataset}
We conducted training and testing on the FLSea dataset\cite{randall2023flsea}, which serves as a challenging benchmark for underwater depth estimation. This dataset comprises two types of underwater visual data collected using different setups: FLSea-stereo, obtained from a stereo configuration, and FLSea-VI, acquired through a monocular visual-inertial setup. The datasets are equipped with either stereo image pairs or monocular images and inertial measurements, scaled depth maps, intrinsic and extrinsic calibrations, enhanced images, and camera poses. Due to variations in the sampled location, significant disparities exist in the oceanic settings between FLSea-VI and FLSea-stereo datasets, resulting in distinct characteristics of underwater imaging.

We primarily utilized the FLSea-VI dataset for training and testing purposes. It comprises $12$ scenes, totaling $22,451$ images with a resolution of $968\times 608$ pixels and captured at a frequency of $10$ Hz. For each scene, we designated  the first $300$ images as candidates for the test set, images $301 \sim 350$ as the validation set, and the remaining images were allocated to the training set. To ensure diversity in the test set scenes, we consistently sampled one image every six from the candidate images to serve as representatives in our test set. Ultimately, we obtained a total of $600$ images for both the test and validation sets, while retaining $18,251$ images exclusively for training purposes. Our proposed split method is referred to as the \texttt{OUC\_split}.

The FLSea-stereo dataset is only used for generalization tests. It consists of four scenes. We followed the same method used in FLSea-VI, where we selected the first $300$ left-view images from each scene as candidates for the test set. We then sampled every sixth image to obtain a set of $200$ test images.

\subsection{Implementation Details and Experimental Setting}
The proposed method is implemented using PyTorch and trained on a single NVIDIA RTX 3090 GPU. The model undergoes training for $50$ epochs on the FLSea-VI dataset, utilizing the \texttt{OUC\_split}. The input resolution is set to $448\times 288$, with a batch size of $28$. AdamW optimizer is employed with a weight decay of $1e-2$. The initial learning rate is set to $5e-4$, following a cosine learning rate schedule. Additionally, drop-path regularization technique is utilized to mitigate overfitting concerns during training. Training the model requires approximately $10$ hours.

\subsection{Contrastive Experiment for Depth Estimation}
We compared our model with a series of monocular depth estimation methods, which can be categorized into two groups: the first group comprises methods based on underwater imaging physical models, including DCP\cite{he2010single}, UDCP\cite{drews2016underwater}, and ULAP\cite{song2018rapid}; the second group consists of DL-baesd methods, including UW-Net\cite{gupta2019unsupervised}, Monodepth2\cite{godard2019digging}, HR-Depth\cite{lyu2021hr}, DIFFNet\cite{zhou2021self}, ManyDepth\cite{watson2021temporal}, MonoViT\cite{zhao2022monovit}, and Lite-Mono\cite{zhang2023lite}. Among these, UW-Net is a GAN-based underwater unsupervised depth estimation method, while the remaining methods are state-of-the-art in terrestrial self-supervised depth estimation. For UW-Net, we directly employed the pretrained model provided by the authors; for the other self-supervised methods, we trained them on the FLSea-VI dataset using the same \texttt{OUC\_split}.

We use the standard error metrics and accuracy metrics for evaluation. The error metrics include mean absolute relative error (Abs Rel), square relative error (Sq Rel), root mean squared error (RMSE) and root mean square logarithmic error (RMSE log). The accuracy metrics include accuracy under threshold ($\delta_i<1.25^i,\mathrm{i=1,2,3}$). Given self-supervised monocular depth estimation methods predict relative depths, following previous works\cite{garg2016unsupervised, zhou2017unsupervised, godard2019digging}, we use median scaling before evaluation to compensate the ambiguity of global scale, which is multiplying the predicted depth maps by the median value of the ground truth. Considering the incomplete depth ground truth, we only use the pixels with effective depth labels for evaluation.

\begin{figure*}[t]
\centering
	\includegraphics[width=0.95\linewidth]{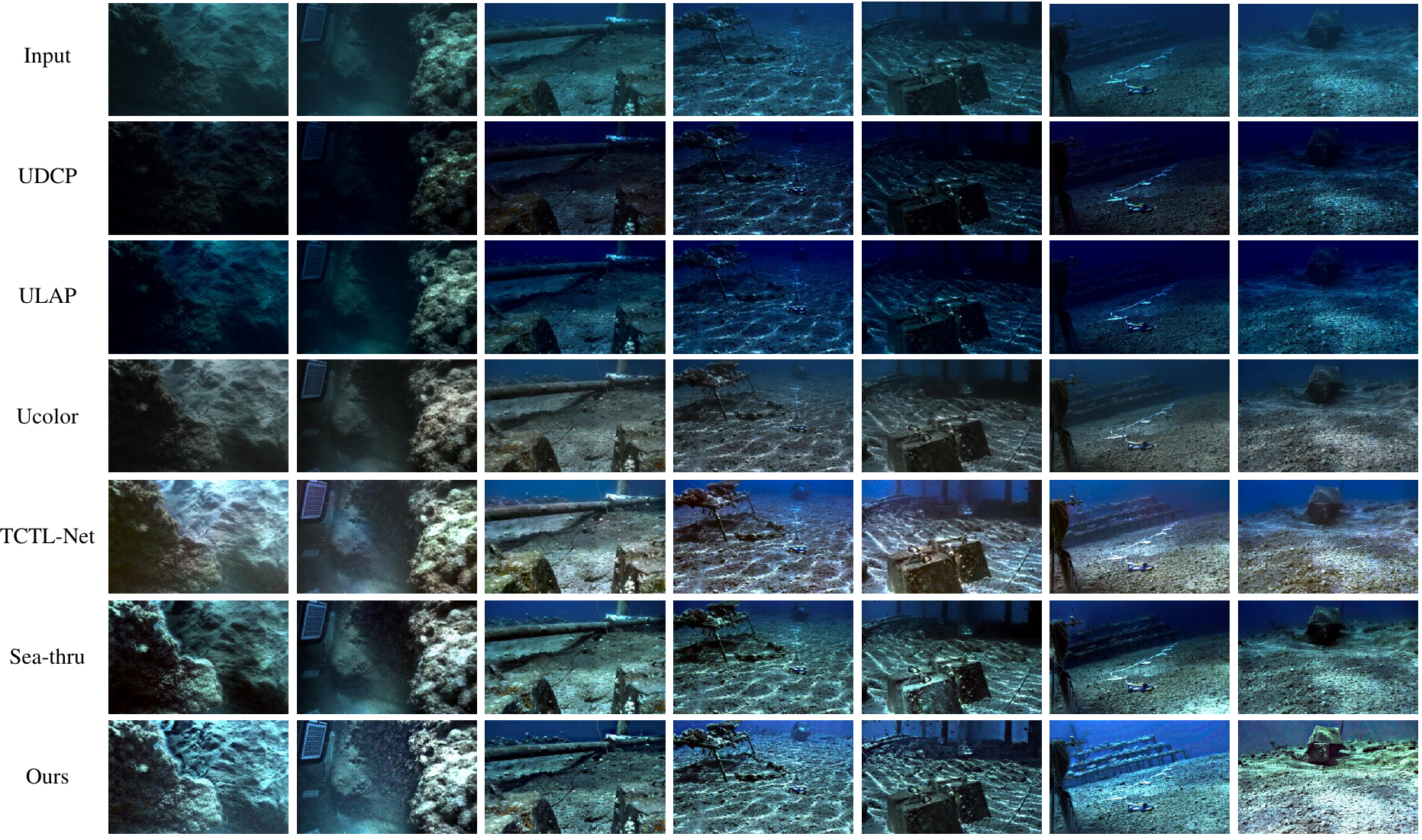}
    \caption{The qualitative comparison of different underwater image enhancement algorithm. The first row comprises the input images, while the second to seventh rows present the outcomes obtained from UDCP\cite{drews2016underwater}, ULAP\cite{song2018rapid}, Ucolor\cite{li2021underwater}, TCTL-Net\cite{li2023tctl}, Sea-thru\cite{akkaynak2019sea} and our proposed method.}
    \label{fig:enhance_imgs}
    \vspace{-1em}
\end{figure*}

The quantitative results are presented in Table~\ref{tab:contrastive}. In general, terrestrial unsupervised monocular depth methods demonstrate a certain degree of robustness to underwater environments and outperform both GAN-based and physics-based methods across all metrics. Among unsupervised/self-supervised depth estimation approaches, the proposed WaterMono not only has the smallest parameter count but also achieves the best performance across all evaluation metrics. WaterMono achieves impressive performance improvements compared to the state-of-the-art method Lite-Mono, without increasing model complexity, by enhancing the learning reliability of key modules and introducing visual enhancement-assisted depth estimation.

The visualized results of all competitors are illustrated in Fig. \ref{fig:test_samples}. It can be observed that methods based on physical models heavily rely on delicate prior assumptions, often resulting in frequent misestimation of depth change trends. For example, in the second column, both DCP and UDCP incorrectly infer the bottom of the rock as a distant object. Moreover, it is evident that certain scenes pose challenges to these approaches in distinguishing foreground from background (refer to the 5th column). The GAN-based UW-Net exhibits significant advancements compared to methods based on physical model, particularly in accurately estimating the overall trend of depth changes. However, due to its training on indoor datasets and synthetic data, it consistently incorporates features that resemble depth maps of indoor scenes into its depth predictions. Monodepth2 more accurately predicts the overall trend of depth changes, but it performs poorly in low-light conditions (refer to the 1st column) and often misestimates backgrounds (refer to the 3rd and 5th columns). HR-Depth, DIFFNet, and ManyDepth exhibit similar overall results to Monodepth2, showing reasonable depth structures but lacking fine local details such as sharp map edges and accurate seawater background depth prediction. MonoViT and Lite-Mono achieve higher levels of edge detail and better performance in low-light environments by introducing Transformers, but they still perform poorly in background regions.

Compared to the aforementioned methods, our results achieve more accurate depth estimation than all other methods, even robustly predicting depth in low-light environments (such as Columns 3 and 5). Moreover, our method generates more precise edges and consistently outperforms in background regions (such as Columns 1, 3, and 7).

\begin{table}[h]
\caption{Quantitative results of image enhancement on the FLSea dataset.\textnormal{All metrics are no-reference metrics and larger values indicate better performance. The best results in each part are in \textbf{bold}. “†” means using a pre-trained model.}}
\fontsize{7.5pt}{11pt}\selectfont
\centering
\begin{tabular}{c|c|c|c|c}
\Xhline{1pt}
Method   & UIQM($\uparrow$)   & UCIQE($\uparrow$) & CPBD($\uparrow$)  & FDUM($\uparrow$)  \\ \hline
UDCP \cite{drews2016underwater}    & 0.050               & 0.496            & 0.535              & 0.455 \\
ULAP  \cite{song2018rapid}   & -0.028              & 0.518            & 0.557              & 0.457 \\ \hline
Ucolor† \cite{li2021underwater}  & 0.067               & 0.527            & 0.540              & 0.368 \\
TCTL-Net† \cite{li2023tctl} & 0.669               & \textbf{0.599}   & 0.431              & 0.482 \\ \hline
Sea-thru \cite{akkaynak2019sea} & 0.501               & 0.592            & \textbf{0.563}     & 0.514 \\
IEB (Ours)     & \textbf{1.048}      & 0.598            & 0.544              & \textbf{0.625} \\ 
\Xhline{1pt}
\end{tabular}
\label{tab:enhancement_metircs}
\end{table}

\subsection{Contrastive Experiment for Image Enhancement}
To evaluate the effectiveness of our IEB module in underwater image enhancement, we compared it with other UIE algorithms, including traditional methods UDCP\cite{drews2016underwater} and ULAP\cite{song2018rapid}, deep learning-based algorithms Ucolor\cite{li2021underwater} and TCTL-Net\cite{li2023tctl}, and Sea-thru\cite{akkaynak2019sea}. For the two DL-based methods, we directly utilized their pre-trained models trained on the UIEB dataset \cite{li2019underwater}. For Sea-thru and our method, the depth information input is obtained from our depth estimation model. We simply convert the predicted relative depth into absolute depth by multiplying it by a fixed scale factor.

We applied each of these methods to enhance 600 images in the test set and evaluated their performance using no-reference metrics. The metrics we utilized include UIQM\cite{panetta2015human}, UCIQE\cite{yang2015underwater}, CPBD\cite{narvekar2011no}, and FDUM\cite{yang2021reference}, which aim to assess the quality of enhanced images in a manner similar to human visual perception.

\begin{figure}[t]
  \centering
   \includegraphics[width=1\linewidth]{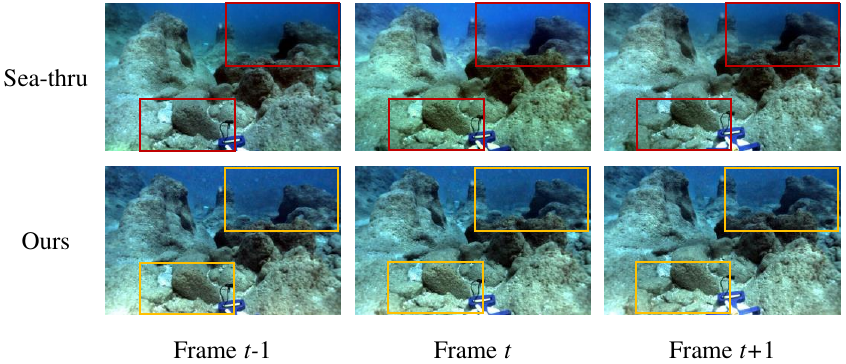}
   \caption{Qualitative testing of inter-frame consistency. The upper images depict the results of three consecutive frames from the FLSea dataset after enhancement with Sea-thru \cite{akkaynak2019sea} and the IEB module, respectively. Compared to Sea-thru, our method exhibits less inconsistencies between frames.}
   \label{fig:enhancement_consistency}
\end{figure}

Fig. \ref{fig:enhance_imgs} provides a visual comparison of different underwater image enhancement methods. As shown in the figure, the original raw underwater images exhibit pronounced blue-green hues, low contrast, blurriness, and low lightness. Both traditional methods UDCP and ULAP perform poorly, exacerbating the blue color cast without significant improvement in visual quality. Deep learning-based methods exhibit more pleasant performance, particularly TCTL-Net, which significantly improves the overall low contrast issue in the images and brings more vibrant colors. However, blurriness in distant areas, which covers details for depth inference, still persists. Finally, in comparison to Sea-thru, we observed that the simplifications implemented in UIFM did not compromise the visual quality of the enhanced images. Instead, they effectively heightened the overall brightness of the images. Furthermore, owing to the incorporation of depth-weighted sharpening, our method exhibits remarkable superiority over others in terms of deblurring, particularly when it comes to eliminating blurriness from distant objects and recovering intricate details concealed. More impressively, as depicted in Fig. \ref{fig:enhancement_consistency}, the simplified UIFM model lead to better inter-frame consistency, which is crucial for self-supervised depth estimation approaches.

\begin{table}[]
\caption{The impact of different enhancement methods on depth estimation performance. \textnormal{The best results in each part are in \textbf{bold}.}}
\scriptsize
\renewcommand{\arraystretch}{1.2}
\centering
\begin{tabular}{c|cccc|c}
\Xhline{1pt}
\multirow{2}{*}{Method} & \multicolumn{4}{c|}{Error($\downarrow$)}                                                                   & Accuracy($\uparrow$) \\ \cline{2-6} 
                        & \multicolumn{1}{l}{Abs Rel} & \multicolumn{1}{l}{Sq Rel} & \multicolumn{1}{l}{RMSE}  & RMSE log & $\delta<1.25$           \\ \hline
None       & \multicolumn{1}{c}{0.116}   & \multicolumn{1}{l}{0.154}  & \multicolumn{1}{l}{1.053} & 0.170    & 0.860          \\
UDCP       & \multicolumn{1}{c}{0.117}   & \multicolumn{1}{l}{0.154}  & \multicolumn{1}{l}{1.069} & 0.174    & 0.854          \\
ULAP       & \multicolumn{1}{c}{0.116}   & \multicolumn{1}{l}{0.156}  & \multicolumn{1}{l}{1.065} & 0.171    & 0.857         \\ 
Ucolor     & \multicolumn{1}{c}{0.115}   & \multicolumn{1}{l}{0.153}  & \multicolumn{1}{l}{1.046} & 0.168    & 0.863         \\ 
TCTL-Net   & \multicolumn{1}{c}{0.113}   & \multicolumn{1}{l}{0.142}  & \multicolumn{1}{l}{1.030} & 0.167    & 0.863         \\ 
Sea-thru   & \multicolumn{1}{c}{0.114}   & \multicolumn{1}{l}{0.146}  & \multicolumn{1}{l}{1.040} & 0.168    & 0.861         \\ 
IEB (Ours)    & \multicolumn{1}{c}{\textbf{0.108}}   & \multicolumn{1}{l}{\textbf{0.132}}  & \multicolumn{1}{l}{\textbf{1.009}} & \textbf{0.161}    & \textbf{0.874}          \\
\Xhline{1pt}
\end{tabular}
\label{tab:enhance_method}
\end{table}

\begin{table}[t]
\caption{Quantitative results of ablation study. \textnormal{The best results in each part are in \textbf{bold}. SD refers to selective distillation, RD refers to rotated distillation and the rotation range $\gamma$ is set to 15°. Full Method means all four components (TGAM, IEB, SD and RD) are enabled.} }
\scriptsize
\renewcommand{\arraystretch}{1.3}
\begin{tabular}{l|cccc|c}
\Xhline{1pt}
\multirow{2}{*}{Method} & \multicolumn{4}{c|}{Error($\downarrow$)}                                                                   & Accuracy($\uparrow$) \\ \cline{2-6} 
                        & \multicolumn{1}{l}{Abs Rel} & \multicolumn{1}{l}{Sq Rel} & \multicolumn{1}{l}{RMSE}  & RMSE log & $\delta<1.25$           \\ \hline
Baseline                & \multicolumn{1}{c}{0.116}   & \multicolumn{1}{l}{0.154}  & \multicolumn{1}{l}{1.053} & 0.170    & 0.860          \\
Baseline+TGAM           & \multicolumn{1}{c}{0.110}   & \multicolumn{1}{l}{0.138}  & \multicolumn{1}{l}{1.018} & 0.164    & 0.872          \\
Baseline+IEB            & \multicolumn{1}{c}{0.108}   & \multicolumn{1}{l}{0.132}  & \multicolumn{1}{l}{1.009} & 0.161    & 0.874          \\
Baseline+SD             & \multicolumn{1}{c}{0.112}   & \multicolumn{1}{l}{0.140}  & \multicolumn{1}{l}{1.012} & 0.164    & 0.870          \\
Baseline+RD             & \multicolumn{1}{c}{0.110}   & \multicolumn{1}{l}{0.146}  & \multicolumn{1}{l}{1.060} & 0.168    & 0.871          \\
Full WaterMono             & \multicolumn{1}{c}{\textbf{0.099}}   & \multicolumn{1}{l}{\textbf{0.117}}  & \multicolumn{1}{l}{\textbf{0.945}} & \textbf{0.152}    & \textbf{0.892}          \\ \Xhline{1pt}
\end{tabular}
\label{tab:ablation_1}
\end{table}

In Table~\ref{tab:enhancement_metircs}, we present the no-reference metrics. In general, the reported scores are consistent with the observed pattern of visual quality variation in Fig. \ref{fig:enhance_imgs}. Our enhancement method achieves the highest UIQM by a remarkable margin over the second place, the second-highest UCIQE, the third-highest CPBD, and the highest FDUM, demonstrating the good visual quality of the enhanced images obtained using our method. 

In Table~\ref{tab:enhance_method}, we further compare the effects of different enhancement methods on self-supervised depth estimation. Generally, better enhancement results lead to larger improvements in depth estimation performance. The two traditional methods provide no assistance for the depth estimation task as they fail to improve the low contrast and blurriness of degraded underwater images. Among the remaining methods, the IEB module, specifically designed for self-supervised depth estimation, exhibits the largest improvement in depth estimation performance. This is because, compared to other methods, it simultaneously addresses the requirements of improving low contrast, reducing blurriness, and maintaining inter-frame consistency, all of which are essential for a rational photometric loss.

\subsection{Ablation Study}
To validate the efficacy of the key components in WaterMono, this section first provides an overview and presents results from the ablation study. Subsequently, more detailed and comprehensive experiments are conducted to demonstrate the ablation outcomes of each component.

\subsubsection{Overall Comparison and Analysis}We first conducted an comprehensive ablation analysis on the FLSea dataset by incorporating the design elements of the proposed WaterMono framework into the baseline model, which served as the teacher network to provide pseudo labels in the second stage. Despite lacking our components, the baseline model still employed the auto-masking (AM) strategy\cite{godard2019digging}, ensuring its competitive performance.  Additionally, the rotation range $\gamma$ for rotated distillation is set to $15^{\circ}$ to match the tilted angles present in the majority of images of the FLSea dataset. The findings presented in Table~\ref{tab:ablation_1} demonstrate that each of our components contributes to enhancing the model's performance, and the comprehensive integration of all components yields the optimal performance.

\subsubsection{Benefits of Teacher-Guided Anomaly Masking (TGAM)} 
\begin{figure}[t]
  \centering
   \includegraphics[width=1\linewidth]{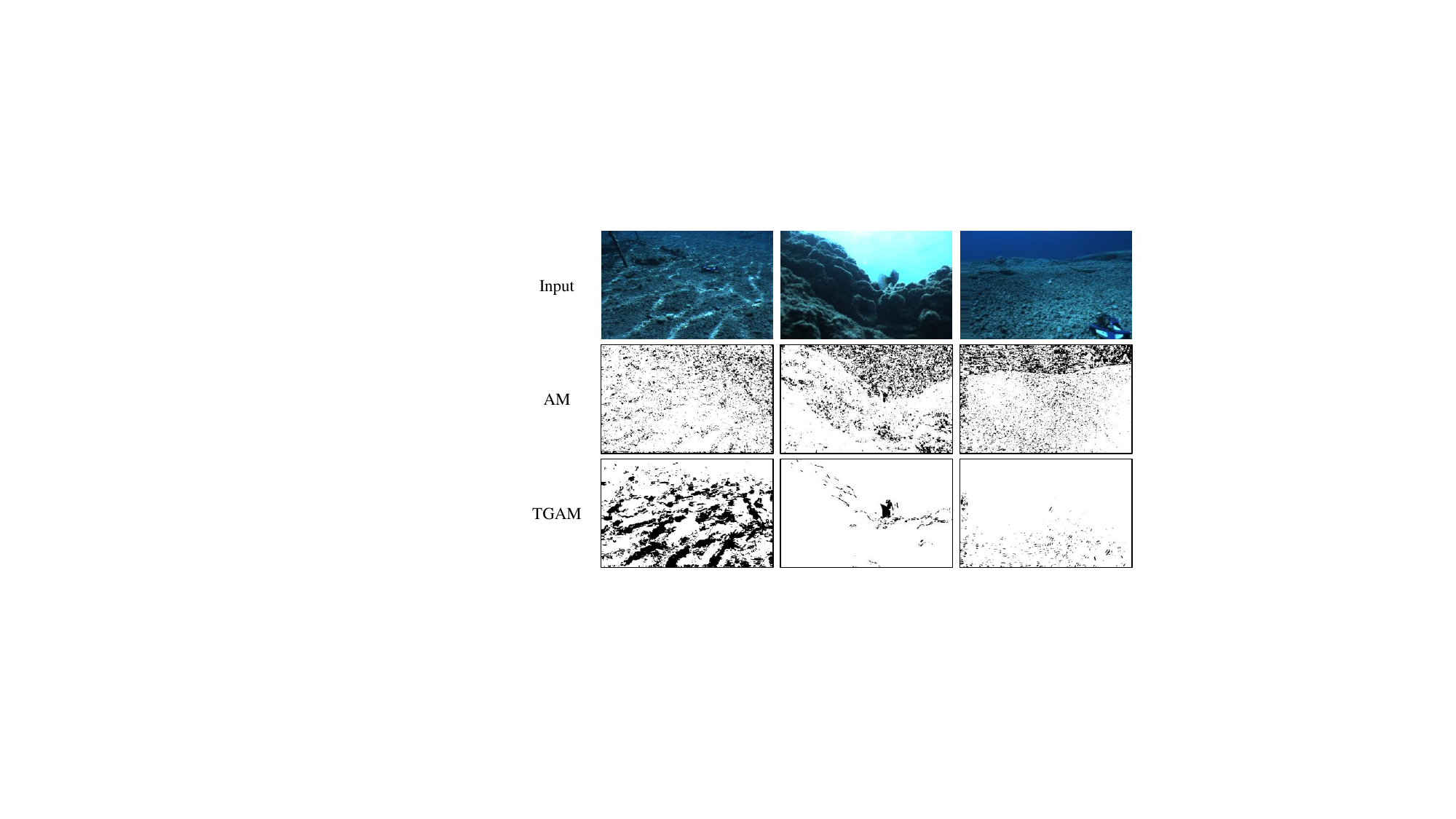}
   \caption{Examples of the masking maps of AM and TGAM.}
   \label{fig:TGAM_AM}
\end{figure}

The experiment regarding the masking strategy employed a baseline model without any masking strategy, which indicates the model's inability to handle anomalous losses. Upon this baseline, we sequentially added AM and TGAM. The results presented in Table~\ref{tab:ablation_2} demonstrate the efficacy of both strategies in enhancing performance by effectively masking anomalous losses. Moreover, simultaneous utilization of these strategies further optimizes overall performance. Fig. \ref{fig:TGAM_AM} illustrates examples of the masking maps of AM and TGAM: AM effectively filters out textureless regions but only partially filters dynamic areas, whereas TGAM filters dynamic regions more comprehensively. This demonstrates the complementary nature of the two masking strategies, which explains why combining them leads to better performance improvement.

\begin{table}[t]
\caption{Ablation study on TGAM.}
\scriptsize
\renewcommand{\arraystretch}{1.3}
\centering
\begin{tabular}{l|cccc|c}
\Xhline{1pt}
\multirow{2}{*}{Method} & \multicolumn{4}{c|}{Error($\downarrow$)}                                                                   & Accuracy($\uparrow$) \\ \cline{2-6} 
                        & \multicolumn{1}{l}{Abs Rel} & \multicolumn{1}{l}{Sq Rel} & \multicolumn{1}{l}{RMSE}  & RMSE log & $\delta<1.25$           \\ \hline
w/o mask       & \multicolumn{1}{c}{0.119}   & \multicolumn{1}{l}{0.164}  & \multicolumn{1}{l}{1.127} & 0.179    & 0.847          \\
w/ AM           & \multicolumn{1}{c}{0.116}   & \multicolumn{1}{l}{0.154}  & \multicolumn{1}{l}{1.053} & 0.170    & 0.860          \\
w/ TGAM           & \multicolumn{1}{c}{0.116}   & \multicolumn{1}{l}{0.153}  & \multicolumn{1}{l}{1.068} & 0.173    & 0.860          \\
w/ TGAM\&AM  & \multicolumn{1}{c}{0.110}   & \multicolumn{1}{l}{0.138}  & \multicolumn{1}{l}{1.018} & 0.164    & 0.872         \\ \Xhline{1pt}
\end{tabular} 
\label{tab:ablation_2}
\end{table}

\begin{table}[t]
\caption{Ablation study on selective distillation.}
\scriptsize
\renewcommand{\arraystretch}{1.3}
\begin{tabular}{l|cccc|c}
\Xhline{1pt}
\multirow{2}{*}{Method} & \multicolumn{4}{c|}{Error($\downarrow$)}                                                                   & Accuracy($\uparrow$) \\ \cline{2-6} 
                        & \multicolumn{1}{l}{Abs Rel} & \multicolumn{1}{l}{Sq Rel} & \multicolumn{1}{l}{RMSE}  & RMSE log & $\delta<1.25$           \\ \hline
w/o distillation       & \multicolumn{1}{c}{0.116}   & \multicolumn{1}{l}{0.154}  & \multicolumn{1}{l}{1.053} & 0.170    & 0.860          \\
distillation           & \multicolumn{1}{c}{0.113}   & \multicolumn{1}{l}{0.154}  & \multicolumn{1}{l}{1.063} & 0.169    & 0.865          \\
selective distillation  & \multicolumn{1}{c}{0.112}   & \multicolumn{1}{l}{0.140}  & \multicolumn{1}{l}{1.012} & 0.164    & 0.870         \\ \Xhline{1pt}
\end{tabular}
\label{tab:ablation_3}
\end{table}

\subsubsection{Benefits of Image Enhancement Boosting (IEB)} Photometric loss has some inherent limitations, one of which is its failure in regions with no texture or low texture. In these areas, regardless of the depth values predicted by the depth estimation network, little to no photometric loss will be generated. The low contrast and blurriness of underwater images significantly diminish texture, creating low-texture regions where photometric loss cannot penalize erroneous depth predictions in degraded areas effectively. Our IEB module effectively improves the overall visual quality of underwater images and reduces blurriness, making the photometric loss computed on enhanced images more reasonable and better optimizing the depth estimation network. The reported quantitative results in Table \ref{tab:enhance_method} serve as experimental evidence, substantiating the aforementioned statements.

\subsubsection{Benefits of Selective Distillation (SD)} We employ selective distillation based on 3D consistency check to filter out unreliable parts in the pseudo labels, which alleviates the instability in self-supervised training and enhances the accuracy of the pseudo labels. In Table~\ref{tab:ablation_3}, we conduct an ablation study on the 3D consistency check. It can be observed that the student network under non-selective distillation fails to considerably surpass the teacher network because errors in the teacher network are also transmitted to the student. However, selective distillation significantly improves the performance of the student network compared to non-selective distillation. This indicates that using 3D consistency checks indeed filters out unreliable portions of pseudo labels, thereby ensuring that the teacher network does not distill inaccurate depth information to the student network as much as possible.

\begin{figure}[t]
  \centering
   \includegraphics[width=1\linewidth]{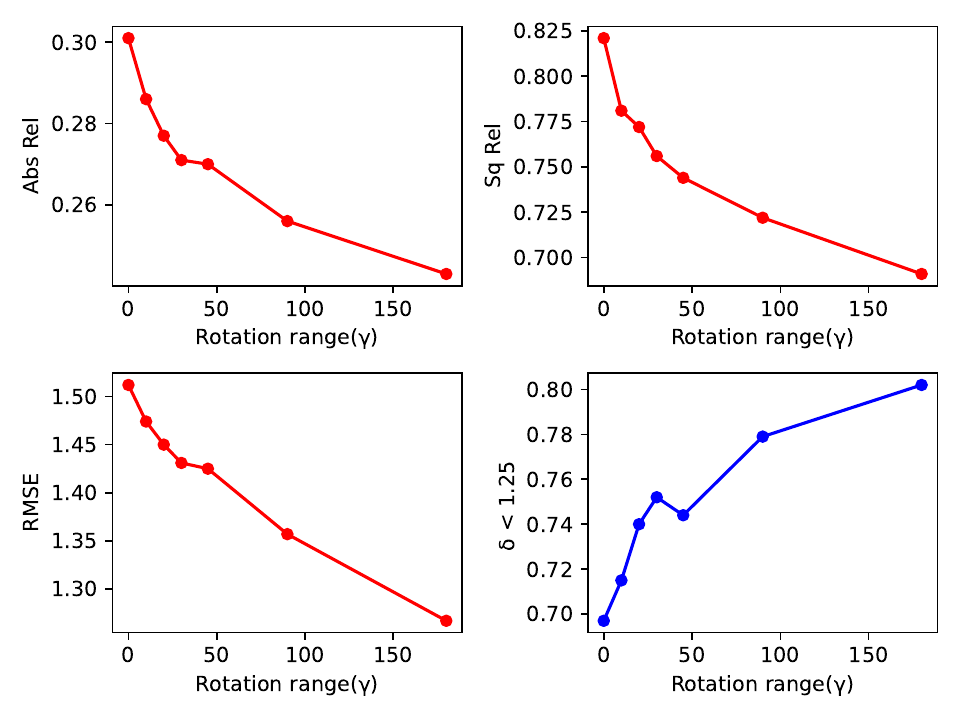}
   \caption{Effect of rotation range $\gamma$. We show the depth performance in the metrics of Abs Rel, Sq Rel, RMSE, and $\delta<1.25$ with different rotation range $\gamma$. The smaller the value, the better in the red line chart, and the worse in the blue.}
   \label{fig:line_plot}
\end{figure}

\begin{figure}[h]
  \centering
   \includegraphics[width=1\linewidth]{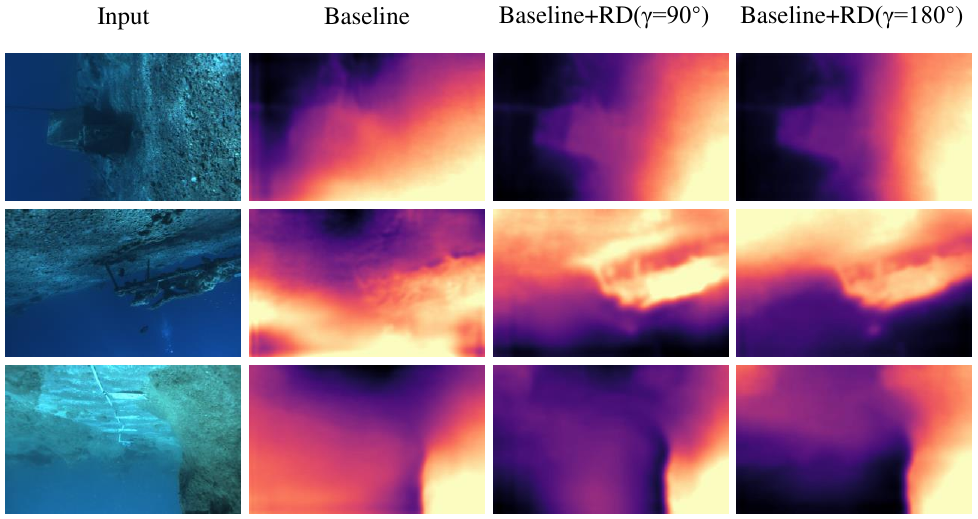}
   \caption{Quantitative results on the challenge test set.}
   \label{fig:rotation_test}
\end{figure}

\subsubsection{Benefits of Rotated Distillation (RD)} Considering that the majority of images within the FLSea dataset are captured from a frontal perspective, in order to more effectively evaluate the model's rotational resilience, we have artificially generated a challenging test set. This particular test set comprises $60$ images; with the first ten being originally captured from rotated camera angles within the FLSea dataset. Images 11 through 50 have been rotated between $30^{\circ}$ - $60^{\circ}$, while images 51 through 60 have been subjected to extreme rotation angles such as those at either $90^{\circ}$ or even up to $180^{\circ}$.

\begin{table}[t]
\caption{Quantitative results on the FLSea-stereo dataset.\textnormal{The best results in each part are in \textbf{bold}.}}
\scriptsize
\renewcommand{\arraystretch}{1.3}
\centering
\begin{tabular}{c|cccc|c}
\Xhline{1pt}
\multirow{2}{*}{Method} & \multicolumn{4}{c|}{Error($\downarrow$)}                                                                   & Accuracy($\uparrow$) \\ \cline{2-6} 
                        & \multicolumn{1}{l}{Abs Rel} & \multicolumn{1}{l}{Sq Rel} & \multicolumn{1}{l}{RMSE}  & RMSE log & $\delta<1.25$           \\ \hline
Monodepth2 \cite{godard2019digging}      & \multicolumn{1}{c}{0.212}   & \multicolumn{1}{l}{0.476}  & \multicolumn{1}{l}{1.911} & 0.297    & 0.642         \\
HR-Depth \cite{lyu2021hr}   & \multicolumn{1}{c}{0.224}   & \multicolumn{1}{l}{0.467}  & \multicolumn{1}{l}{1.768} & 0.313    & 0.624          \\
DIFFNet  \cite{zhou2021self}   & \multicolumn{1}{c}{0.225}   & \multicolumn{1}{l}{0.579}  & \multicolumn{1}{l}{2.265} & 0.356    & 0.609          \\
MonoViT \cite{zhao2022monovit} & \multicolumn{1}{c}{0.227}   & \multicolumn{1}{l}{0.463}  & \multicolumn{1}{l}{1.882} & 0.308    & 0.611                      \\
Lite-Mono \cite{zhang2023lite} & \multicolumn{1}{c}{0.226}   & \multicolumn{1}{l}{0.444}  & \multicolumn{1}{l}{1.765} & 0.290    & 0.607                      \\
WaterMono (Ours)      & \multicolumn{1}{c}{\textbf{0.209}}   & \multicolumn{1}{l}{\textbf{0.397}}  & \multicolumn{1}{l}{\textbf{1.616}} & \textbf{0.266}    & \textbf{0.667}          \\ \Xhline{1pt}
\end{tabular}
\label{tab:generalization}
\end{table}

On the challenging test set, we evaluated the performance of the model using metrics including Abs Rel, Sq Rel, RMSE, and $\delta<1.25$ for different rotation range parameter -- $\gamma$. The visual results are presented in Fig. \ref{fig:line_plot}. It is evident that, overall, the model's performance on the challenge test set consistently improves with an increasing rotation range $\gamma$. This suggests a positive correlation between larger values of $\gamma$ and enhanced rotational robustness of the model.

Besides, it is noteworthy that we observed a significant risk of overfitting in self-supervised depth estimation to the vertical position of the image. Specifically, like other self-supervised depth estimation frameworks, the baseline model tends to predict larger depths for the upper parts of the images and smaller depths for the lower parts, as shown in the second column of Fig. \ref{fig:rotation_test}. While this may not be problematic on land where cases violating this pattern are rare, underwater environments pose a different scenario. AUVs may navigate at extreme tilt angles, even in upside-down orientations, where erroneous depth estimations can lead to severe consequences.

We observed that when the rotation range $\gamma$ exceeds $90^{\circ}$, there is a certain probability of breaking the relationship between depth and the vertical position of the image, which forces the model to capture more effective depth cues such as semantics. This implies that, as depicted in Fig. \ref{fig:rotation_test}, even under extreme underwater conditions with inverted orientations, models employing rotated distillation can accurately predict depth, while the predictions of the baseline (Lite-Mono's results) are entirely erroneous in such scenarios.

\subsection{Generalization Evaluation}
\begin{figure}[t]
  \centering
   \includegraphics[width=1\linewidth]{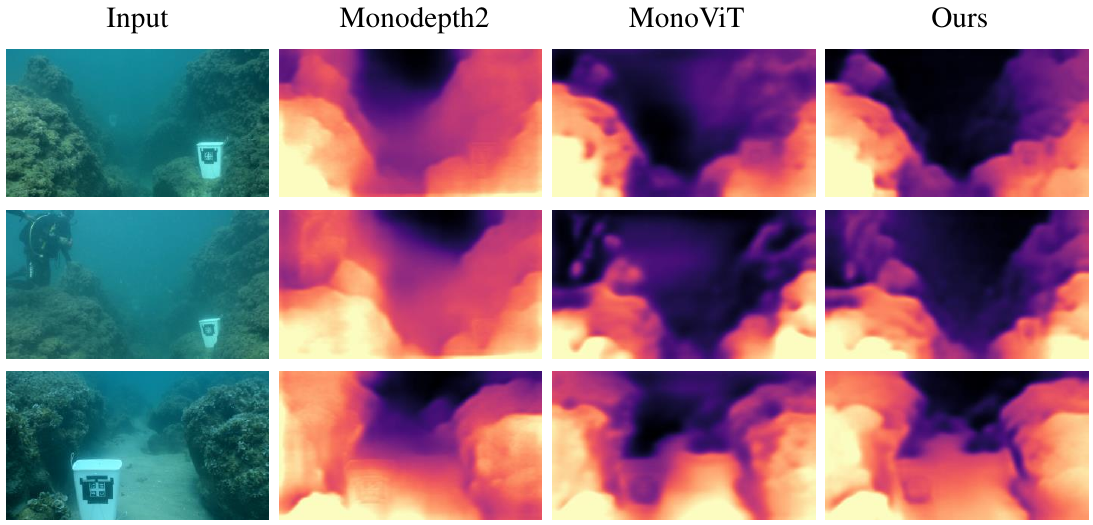}
   \caption{Quantitative results on the FLSea-stereo dataset. All methods were trained
on FLSea-VI using \texttt{OUC\_split}.}
   \label{fig:generalization_test}
\end{figure}

To further assess the generalization capability of the proposed WaterMono, we employed the model trained on the FLSea-VI dataset to evaluate the FLSea-stereo dataset without any fine-tuning. In comparison to FLSea-VI, FLSea-stereo exhibits significantly murkier conditions in its water bodies and encompasses numerous homogeneous sandy areas and rocky outcrops adorned with aquatic vegetation, which are absent in FLSea-VI. The performance of our method is presented in Table~\ref{tab:generalization}, where a comparison with five other deep learning-based approaches demonstrates that our method outperforms them all and exhibits superior generalization capability. Fig. \ref{fig:generalization_test} showcases examples of depth estimation results.

\section{conclusion}
This paper aims to achieve accurate self-supervised monocular depth estimation in underwater environments. We analyze the challenges posed by this unique setting and propose several techniques to address them, including TGAM for dynamic region filtering, depth-guided underwater image enhancement for depth estimation boosting, and rotated distillation for handling diverse camera angles. Our framework leverages the intrinsic connections between underwater depth estimation and image enhancement, enabling simultaneous acquisition of precise depth estimates and visually appealing enhanced images. Extensive experiments demonstrate that WaterMono advances the state-of-the-art in underwater monocular depth estimation. We believe our work can greatly contribute to vision-based navigation and decision-making in AUVs.

However, our framework also has limitations primarily due to the introduction of additional training steps in the two-stage training process. In the future, we will explore using self-distillation to compress these training steps.

\ifCLASSOPTIONcaptionsoff
  \newpage
\fi

\bibliographystyle{IEEEtran}

\bibliography{reference}


\end{document}